\documentclass[letterpaper]{article} 
\usepackage{aaai2027}  
\nocopyright

\usepackage[hyphens]{url}  
\usepackage{graphicx} 
\usepackage{natbib}  
\usepackage{caption} 
\usepackage{amsmath,amssymb}
\usepackage{booktabs}

\newcommand{\method}{AERA}
\newcommand{\aec}{AEC}

\title{AERA: Adaptive Evidence Residual Allocation for Efficient Test-Time Reasoning}

\author{
    Ziming Wang\textsuperscript{\rm 1},
    Ivor Tsang\textsuperscript{\rm 2},
    Hangwei Qian\textsuperscript{\rm 2,3}\thanks{Corresponding author. Hangwei Qian}
}

\affiliations{
    \textsuperscript{\rm 1}National University of Singapore, Singapore\\
    \textsuperscript{\rm 2}Agency for Science, Technology and Research (A*STAR), Centre for Frontier AI Research (CFAR), Singapore\\
    \textsuperscript{\rm 3}Nanyang Technological University, Singapore
}

\begin{document}

\maketitle

\begin{abstract}
Test-time scaling improves language-model reasoning by generating additional candidate solutions, but allocating the same inference budget to every problem is computationally wasteful. Existing adaptive stopping methods commonly rely on confidence, agreement, or answer stability, implicitly assuming that stronger current evidence indicates that further computation is unnecessary. We show that this assumption can fail: checkpoint-level correctness evolves non-monotonically, and observable evidence may strengthen before an answer collapses or weaken before it recovers. Motivated by this mismatch, we introduce Adaptive Evidence Residual Allocation (AERA), a sequential controller that learns whether additional computation is likely to recover a better answer from checkpoint-observable evidence. AERA characterizes cumulative response prefixes using answer-distribution, temporal, re-solving, semantic, and compute features, and repeatedly decides whether to stop or allocate the next response block. Future checkpoint correctness is used only to construct offline supervision and is never available to the controller at inference time. Across GSM8K and GPQA Diamond, AERA identifies question-specific residual opportunities while substantially reducing inference computation. In a frozen-threshold incremental-generation evaluation on 300 untouched GSM8K questions, AERA achieves 92.61\% accuracy versus 93.01\% with 128 responses while reducing completion tokens by 95.99\%. These results suggest that adaptive reasoning should estimate the future value of computation rather than equating present confidence with correctness.
\end{abstract}

\section{Introduction}

Large reasoning models improve problem solving by spending additional
computation at test time, either through longer traces, multiple candidate
solutions, or renewed attempts after an unproductive path
\cite{snell2024scaling,manvi2024adaptive,wang2026re2}.
The gains are substantial, but uniform scaling is wasteful: a fixed budget
continues sampling both questions resolved after a few responses and questions
whose answers remain unsettled.
The central problem in adaptive inference is therefore not simply how to reduce
computation, but how to determine when another unit of reasoning is 
useful.

\begin{figure}[!t]
\centering
\includegraphics[
  width=\columnwidth,
  keepaspectratio
]{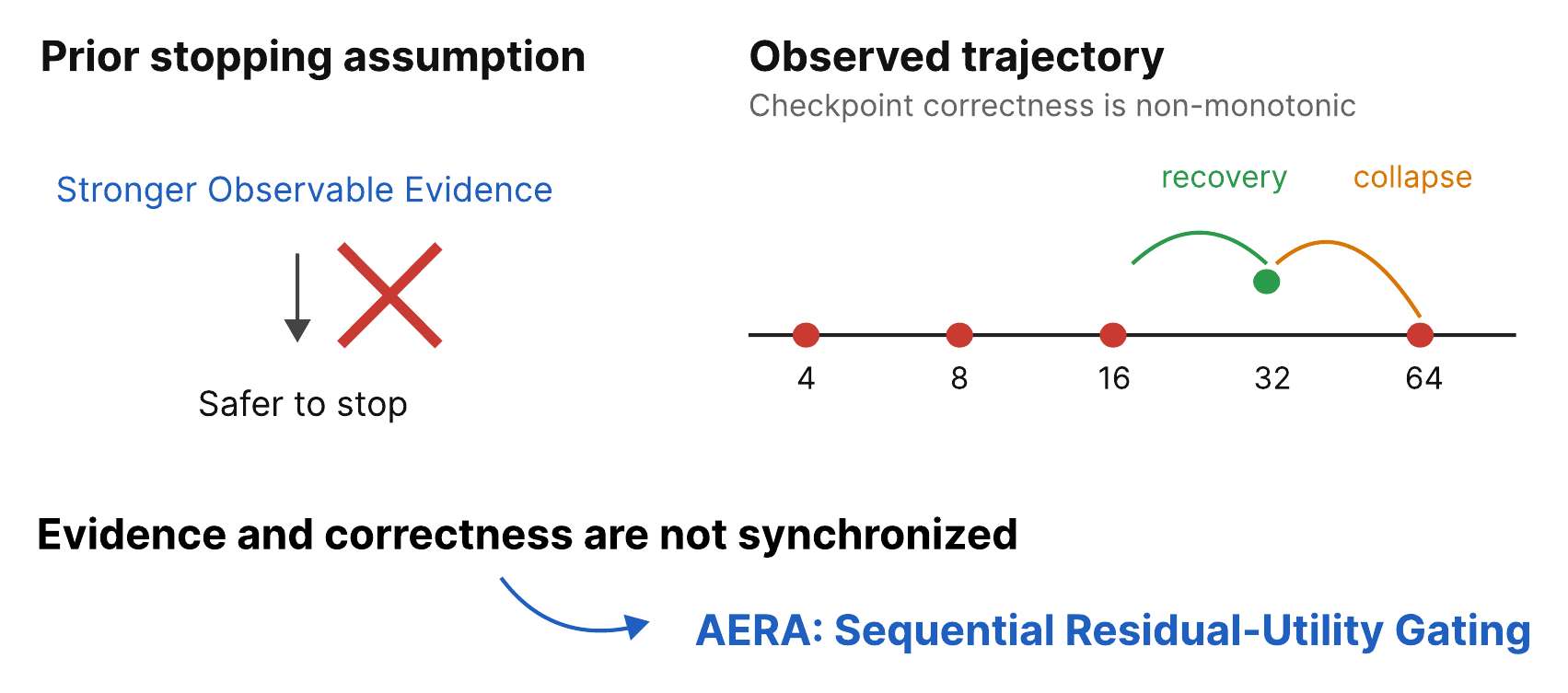}
\caption{Motivation for \method{}. A representative GPQA trajectory first
recovers and then collapses, contradicting the assumption that stronger
observable evidence always makes stopping safer. This mismatch motivates
sequential residual-utility control.}
\label{fig:intro-trajectory}
\end{figure}

Recent work has progressively moved from fixed scaling toward adaptive
reasoning.
Training-based approaches learn query-dependent inference budgets or decide
whether explicit thinking is needed
\cite{yu2025ibpo,tu2025autothink}.
At the response level, Early-Stopping and Difficulty-Adaptive
Self-Consistency vary the number of sampled solutions using answer stability
and estimated question difficulty \cite{li2024esc,wang2024dsc}.
Within a single reasoning trace, DEER and confidence-guided compression monitor
intermediate signals to terminate or shorten generation
\cite{yang2025deer,qiao2025concise}.
Together, these developments establish adaptive computation as a promising
route to efficient reasoning, but leave a fundamental question unresolved:
what should a stopping criterion estimate when reasoning quality evolves
non-monotonically?

Most existing methods base their stopping decisions on evidence observed at
the current step.
They terminate computation when confidence is high, entropy is low, or sampled
answers agree.
Such signals describe the current response population, but they do not directly
measure the value of computation that has not yet been performed.
Conflating the two is harmless only if stronger evidence reliably accompanies
correctness and additional reasoning improves answers monotonically.

\begin{figure}[!t]
\centering
\includegraphics[width=\columnwidth]{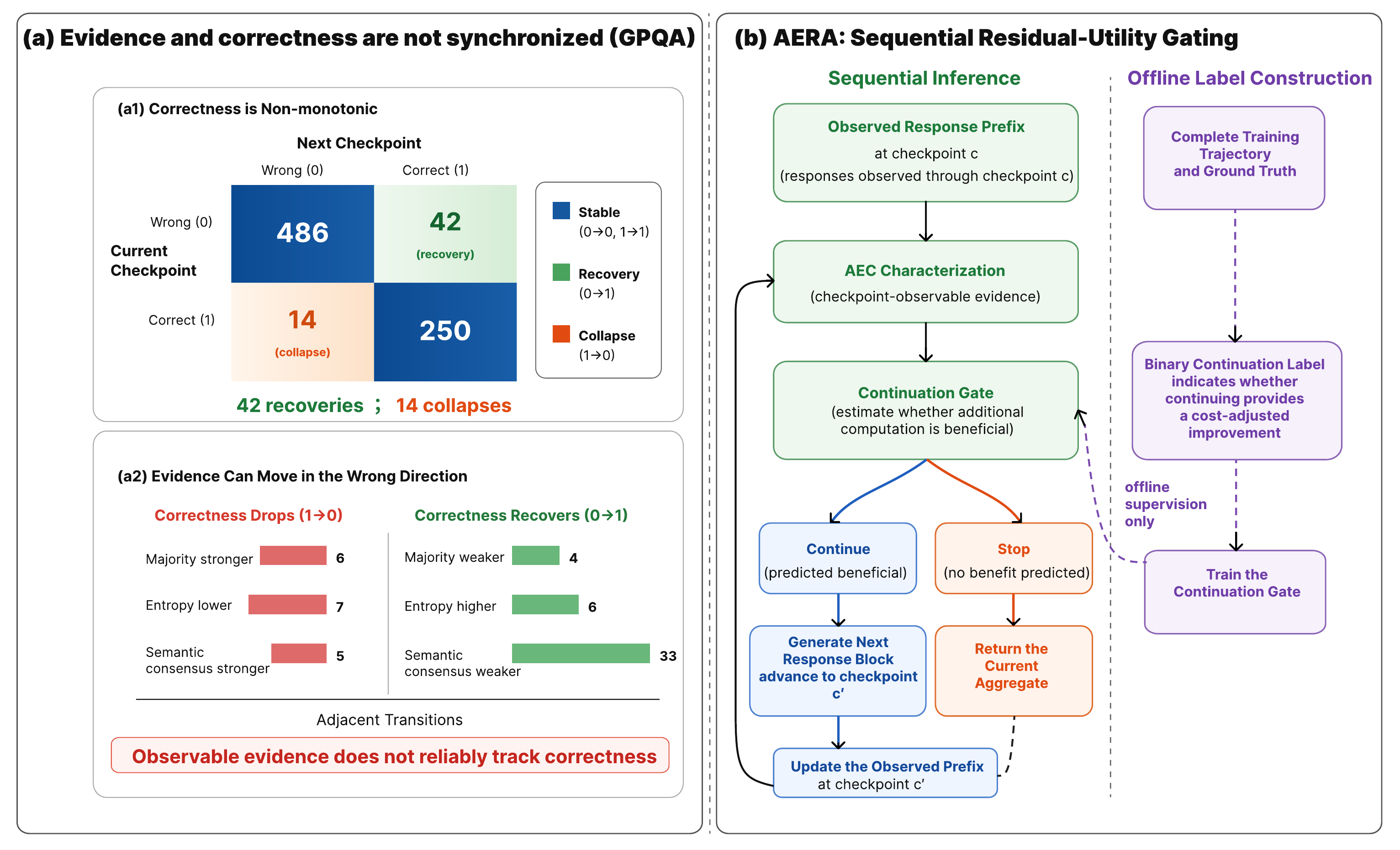}
\caption{Empirical motivation and overview of \method{}. Observable evidence
can move opposite to correctness: evidence sometimes
strengthens during collapse and weakens during recovery. \method{} therefore
characterizes each observed prefix and sequentially estimates whether the next
checkpoint remains beneficial.}
\label{fig:overview}
\end{figure}

Reasoning trajectories violate this assumption.
Prior work similarly finds that accuracy can decrease once a reasoning chain
grows beyond a task- and model-dependent length \cite{wu2026whenmore}.
Across 792 adjacent GPQA checkpoint transitions, we observe 42 recoveries from
an incorrect to a correct aggregate and 14 collapses in the opposite direction.
Figure~\ref{fig:intro-trajectory} illustrates a trajectory that first recovers
and then collapses.
The aggregate transition statistics in Figure~\ref{fig:overview} further show
that conventional evidence signals sometimes strengthen during collapse and
weaken during recovery.
Evidence is therefore informative about the observed state without being
equivalent to current correctness---or to the prospect of future recovery.

This mismatch suggests a different decision target.
Rather than asking whether the current answer appears correct, an adaptive
controller should estimate whether allocating additional computation can
improve it
enough to justify the cost.
We call this quantity \emph{residual utility}.
Its supervision may be constructed retrospectively from complete training
trajectories, while the resulting policy must act prospectively from the
response prefix available at inference.
This separation turns future correctness into a training signal rather than a
privileged controller input.

We operationalize this principle with \method{}
(Figure~\ref{fig:overview}).
At each checkpoint, Adaptive Evidence Characterization (\aec{}) summarizes the
observed response prefix using features that capture answer concentration,
temporal dynamics, re-solving behavior, semantic agreement, and computation
already consumed.
A checkpoint-shared residual-utility gate then determines whether to allocate
the next block of responses.
If computation continues, \method{} updates its evidence state from the
expanded response set and reevaluates the decision at the next checkpoint,
rather than committing in advance to a fixed terminal budget.

This work makes three contributions.
First, we identify and quantify the mismatch between observable evidence and
checkpoint correctness, including recovery, collapse, oscillation, and
confidence changes in the wrong direction.
Second, we formulate adaptive allocation as residual-utility prediction and
introduce \method{}, which separates checkpoint-observable evidence from future
correctness used only for supervision.
Third, we evaluate this formulation with full offline frontiers, nested
calibration, and frozen-threshold generation, obtaining a near-lossless GSM8K
operating point selected without test access.

\section{Related Work}

\paragraph{Controlling reasoning before inference.}
One family of efficient-reasoning methods changes the reasoner or specifies its
budget before intermediate evidence is observed
\cite{liu2025efficientsurvey}.
Post-training can teach a model when to switch between direct answering and
explicit thinking
\cite{zhang2025adaptthink,jiang2025hybrid,tu2025autothink}, learn
inference-budget-aware policies \cite{yu2025ibpo}, compress the
length of learned reasoning traces \cite{ma2025cotvalve}, optimize for concise
solutions \cite{dumitru2025conciserl}, or calibrate chain-of-thought length
during reinforcement learning \cite{hu2026smartthinker}.
Structural optimization and adaptive parallel reasoning provide complementary
ways to reorganize the computation itself
\cite{gui2026cosmo,pan2025apr}.
Training-free alternatives impose token budgets, elicit compact intermediate
work, or bypass explicit reasoning
\cite{han2024tokenbudget,muennighoff2025s1,xu2025chainofdraft,ma2025nothinking}.
These methods reduce the cost of producing a reasoning trace, whereas
\method{} leaves the reasoner unchanged and allocates additional sampled
responses after observing their aggregate state.

\paragraph{Adaptive stopping during inference.}
Response-level methods are the closest setting to ours.
Adaptive-Consistency uses a Beta posterior over the two leading answer counts
to stop once the current leader is sufficiently stable
\cite{aggarwal2023adaptive}.
Early-Stopping Self-Consistency terminates sampling after the aggregated answer
becomes stable \cite{li2024esc}; Difficulty-Adaptive Self-Consistency combines
question difficulty with posterior evidence to assign sample budgets
\cite{wang2024dsc}; and answer-convergence methods stop when intermediate
answers reach a stable decision \cite{liu2025answerconvergence}.
A parallel line controls individual reasoning traces.
DEER probes trial answers at reasoning-transition points
\cite{yang2025deer}, SEAL steers latent reasoning representations
\cite{chen2025seal}, and confidence-guided methods use certainty, entropy, or
hidden-state verification signals to curtail redundant computation
\cite{fu2025certainty,qiao2025concise,zhang2025selfverify}.
Manvi et al.\ predict whether restarting or extending an individual generation
can improve its answer \cite{manvi2024adaptive}. In contrast, \method{} acts on
cumulative response populations at aligned checkpoints, uses future canonical
aggregate correctness only to supervise continuation, and re-observes the
population after each block. Unlike the present-convergence signals of ESC and
ASC, \method{} predicts future recoverability from observable evidence when
convergence and correctness may diverge. Related controllers use prefix value
or uncertainty to trigger continuation or abstention
\cite{davidov2026quit,ding2025adanav}.

\paragraph{Non-monotonic reasoning and residual utility.}
Test-time scaling and aggregation improve robustness on average
\cite{snell2024scaling,manvi2024adaptive}, but individual trajectories need not
improve monotonically.
Indeed, reasoning accuracy can follow an inverted-U relationship with
chain-of-thought length \cite{wu2026whenmore}.
Re$^2$ learns to revisit unproductive reasoning paths \cite{wang2026re2},
negative final trajectories can retain useful intermediate states
\cite{tian2026glow}, and step-level analyses expose failures and recoveries
within long reasoning processes \cite{xu2026stepflow}.
\section{Empirical Motivation: Evidence Is Not Correctness}
\label{sec:phenomenon}

\paragraph{Trajectory construction.}
To test whether observable confidence is a reliable stopping target, we analyze
cumulative response prefixes at checkpoints 4, 8, 16, 32, 64, and 128.
At each checkpoint, the original Re$^2$ aggregation and answer-normalization
procedure selects an answer, whose ground-truth correctness is denoted by
$z_{q,c}\in\{0,1\}$.
Adjacent checkpoints are categorized as stable wrong
($0\!\rightarrow\!0$), recovery ($0\!\rightarrow\!1$), collapse
($1\!\rightarrow\!0$), or stable correct ($1\!\rightarrow\!1$).
We call a trajectory oscillatory when its correctness changes at least twice.
This analysis uses correctness only to characterize the phenomenon; it is not a
controller input.

\subsection{Correctness Evolves Non-Monotonically}

Table~\ref{tab:gpqa-transitions} reports the GPQA analysis for checkpoints 4 through 64, the range for which aligned \aec{} states are available.
Recovery occurs three times as often as collapse (42 versus 14 transitions), directly invalidating both ``more compute cannot repair an error'' and ``more compute cannot harm a correct answer.''
On GSM8K, the six-checkpoint correctness trajectories contain 33 monotonic recoveries, 11 oscillatory recovered cases, one monotonic collapse, and six oscillatory failures among 1,319 questions.

\begin{table}[t]
\centering
\small
\begin{tabular}{lrrrr}
\toprule
Transition & $N$ & $\Delta$ majority & $\Delta$ entropy & $\Delta$ semantic \\
\midrule
$0\!\rightarrow\!0$ & 486 & $+0.050$ & $+0.044$ & $-0.005$\\
$0\!\rightarrow\!1$ & 42  & $+0.755$ & $+0.073$ & $-0.003$\\
$1\!\rightarrow\!0$ & 14  & $-0.022$ & $+0.118$ & $-0.003$\\
$1\!\rightarrow\!1$ & 250 & $-0.029$ & $+0.069$ & $-0.004$\\
\bottomrule
\end{tabular}
\caption{Adjacent GPQA correctness transitions over aligned \aec{} states (4$\rightarrow$8 through 32$\rightarrow$64). Entries report transition counts and mean feature changes; entropy denotes normalized entropy.}
\label{tab:gpqa-transitions}
\end{table}

\subsection{Evidence Can Move Against Correctness}

\begin{figure}[t]
\centering
\includegraphics[width=\columnwidth]{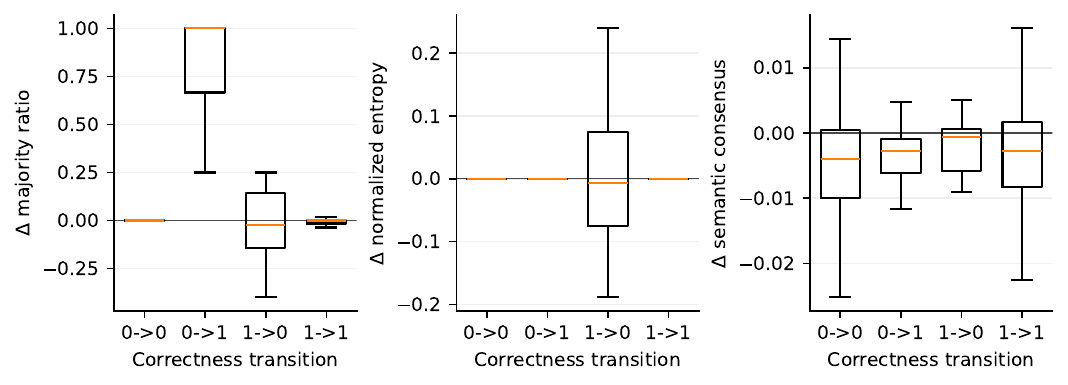}
\caption{Changes in observable evidence grouped by adjacent GPQA correctness transition. Boxes summarize the distributions and the horizontal line marks zero. Evidence coordinates overlap substantially across correctness outcomes.}
\label{fig:transition-deltas}
\end{figure}

Aggregate means alone can hide the counterexamples that motivate our method.
Among the 14 GPQA collapses, six show a larger majority ratio, seven show lower entropy, and five show higher semantic consensus; one improves all three evidence proxies while becoming incorrect.
Conversely, four recoveries occur while majority ratio falls, six while entropy rises, and 33 while semantic consensus falls.
These cases establish an empirical distinction, not a causal claim: individual evidence coordinates are informative but cannot be treated as correctness labels.
Figure~\ref{fig:transition-deltas} shows that their transition-conditioned distributions substantially overlap.

\section{\method{}: Adaptive Residual Compute Allocation}

\subsection{Method Overview}

\method{} controls a frozen checkpointed reasoner without changing how individual responses are generated.
The gate is shared across checkpoints and is evaluated again after every
allocated response block.

\subsection{Residual-Utility Objective}

Let $q\in\mathcal{Q}$ denote a question and
$\mathcal{R}_{q,c}=(r_{q,1},\ldots,r_{q,c})$ the prefix containing its
first $c$ sampled responses.
Decisions are made at checkpoints
$\mathcal{C}=\{4,8,16,32,64,128\}$.
A frozen aggregation rule $A(\mathcal{R}_{q,c})$ selects the answer returned
from the current prefix, and
$z_{q,c}=\mathbf{1}\!\left\{A(\mathcal{R}_{q,c})=a_q^\star\right\}$ denotes its
correctness with respect to ground truth $a_q^\star$.

At checkpoint $c$, \aec{} produces an observable state
$s_{q,c}=\phi(q,\mathcal{R}_{q,c},c)$.
The controller chooses between \textsc{Stop}, which returns
$A(\mathcal{R}_{q,c})$, and \textsc{Continue}, which allocates enough additional responses to reach the next checkpoint and then observes a new state.
Let $K_{q,c}$ be cumulative generation cost and $K_{q,\max}$ the cost at checkpoint 128.

We define the utility advantage of continuing to a later checkpoint $t>c$ over stopping now as
\begin{equation}
\Delta U_q(c\!\rightarrow\!t)
= z_{q,t}-z_{q,c}
-\lambda\frac{K_{q,t}-K_{q,c}}{K_{q,\max}},
\label{eq:delta-u}
\end{equation}
where $\lambda$ controls the trade-off between correctness improvement and
normalized incremental generation cost.
We define the binary continuation label as
\begin{equation}
y_{q,c} =
\mathbf{1}\!\left[
\max_{t\in\mathcal{C}:t>c}\Delta U_q(c\!\rightarrow\!t)>\delta
\right],
\label{eq:label}
\end{equation}
with $\delta=0$ in our experiments.
The label equals one when at least one reachable later checkpoint has higher
cost-adjusted correctness than stopping at $c$.
Rather than specifying a terminal budget, it supervises the local decision to
continue from checkpoint $c$.

\paragraph{Recoverability as a special case.}
The general residual-utility definition accommodates real-valued task rewards and costs.
Under the present experiments, however, $z$ is binary, $0\leq(K_{q,t}-K_{q,c})/K_{q,\max}\leq1$, $\delta=0$, and $0\leq\lambda<1$.
In this regime,
\begin{equation}
y_{q,c}=1
\quad\Longleftrightarrow\quad
z_{q,c}=0\ \text{and}\ \exists t>c:\ z_{q,t}=1.
\label{eq:recoverability}
\end{equation}
Indeed, a wrong-to-correct transition has gain
$1-\lambda(K_{q,t}-K_{q,c})/K_{q,\max}>0$, whereas all other binary
transitions have non-positive gain after cost.
Thus, in the evaluated setting, residual-utility prediction reduces exactly to
\emph{recoverability prediction}: the positive states are currently incorrect
but become correct at a later checkpoint.
This equivalence also explains why varying $\lambda$ within the tested range
does not change the binary labels; real-valued rewards would recover the more
general cost-sensitive objective in Equation~\ref{eq:delta-u}.
Accordingly, our experiments validate this binary recoverability special case,
not the full graded-utility formulation.

\paragraph{Sequential consistency.}
For a currently incorrect state at $c$, let $t^\star$ be its earliest future
correct checkpoint. At every scheduled checkpoint $c\leq u<t^\star$, we have $z_{q,u}=0$
and $t^\star$ remains reachable, so Equation~\ref{eq:recoverability} gives
$y_{q,u}=1$. At $t^\star$, $z_{q,t^\star}=1$, and no later binary outcome can
produce a positive correctness gain after nonnegative cost, so
$y_{q,t^\star}=0$. An exact-label policy therefore advances one block at a
time to the earliest recovery and stops there; if the current state is already
correct, it stops immediately. Thus the any-future label is consistent with
the local sequential action under the evaluated binary assumptions, although
prediction errors can still terminate a learned rollout early.

\subsection{Training Labels and Information Boundary}

The controller observes a strictly smaller information set than the procedure
used to construct its labels.
At checkpoint $c$, its available information is
\[
\mathcal{I}_{q,c}
=\{q,\mathcal{R}_{q,1:c},c,K_{q,c},\mathcal{C}\}
\]
all of which is observable when the decision is made.
Ground truth and later checkpoint outcomes are excluded from
$\mathcal{I}_{q,c}$ and used only for offline label construction.
Canonical correctness $z_{q,c}$ is computed by applying the same frozen answer extraction, normalization, and Re$^2$ aggregation rule used for evaluation to the first $c$ responses.
It is never estimated from an evidence feature and never enters $s_{q,c}$.
This prevents majority strength, entropy, or semantic similarity from becoming circular definitions of correctness.

Equation~\ref{eq:recoverability} describes the offline target, while the learned predictor must infer its probability from checkpoint-observable evidence alone.

\subsection{Observable Evidence Characterization}

\aec{} maps the observed response prefix to a numerical description of the current reasoning state.
It deliberately characterizes \emph{what has been observed} without estimating correctness.
The representation combines five complementary groups:
(i) \emph{budget and cost}, including checkpoint fraction, the next scheduled
checkpoint, and cumulative cost observed so far;
(ii) \emph{answer distribution}, including majority ratio, normalized entropy,
unique-answer ratio, and the top-two gap;
(iii) \emph{temporal change}, including changes in majority ratio, entropy, and redo
rate, together with top-answer change and persistence;
(iv) \emph{re-solving behavior}, summarized by redo rate and its change; and
(v) \emph{semantic state}, including consensus, dispersion, similarity statistics,
centroid shift, new-block novelty, and a 32-dimensional projection.
Every component is measurable from the question, the responses observed by checkpoint $c$, or the deterministic checkpoint schedule.
Ground-truth answers, checkpoint correctness, later responses, future costs, and residual labels are not part of the state.

\paragraph{Answer-distribution state.}
Responses are first mapped to normalized candidate answers.
If $n_a$ responses support answer $a$ among $c$ observed responses, \aec{} records the largest empirical mass $\max_a n_a/c$, normalized entropy, the number of distinct normalized answers relative to $c$, and the gap between the two largest masses.
These variables summarize concentration without consulting the ground-truth answer.

\paragraph{Temporal state.}
For adjacent checkpoints, \aec{} records changes in concentration, entropy, and redo rate, whether the top answer changes, and how long it persists.
Temporal features make recovery and destabilization observable without asserting whether either event is correct.
At the first checkpoint, deltas use fixed neutral values rather than information from an unobserved predecessor.

\paragraph{Semantic state.}
Response embeddings define pairwise similarity summaries, dispersion around the current centroid, movement of that centroid, and novelty of the newly generated response block.
The 32 projection coordinates provide a compact representation of the semantic population.
Encoder weights and projection parameters are frozen before gate fitting; no correctness label is used to update the semantic encoder.

\subsection{Residual Gate Learning}

The gate estimates
\[
p_\theta(s_{q,c}) = P_\theta(y_{q,c}=1\mid s_{q,c}).
\]
Our implementation is a two-hidden-layer MLP with dimensions 64 and 32, LayerNorm, GELU activations, and dropout 0.1.
Class-weighted binary cross-entropy addresses the sparse positive labels.
Let $w_1$ and $w_0$ be class weights computed on training questions only.
For compactness, write $p_{q,c}=p_\theta(s_{q,c})$.
The weighted loss for one checkpoint is
\[
\begin{aligned}
\ell_{q,c}={}&-w_1y_{q,c}\log p_{q,c}\\
&-w_0(1-y_{q,c})\log(1-p_{q,c}).
\end{aligned}
\]
The training objective is
\[
\mathcal{L}(\theta)
=\sum_{(q,c)\in\mathcal{D}_{\mathrm{train}}}\ell_{q,c}.
\]
Feature medians, means, and standard deviations are fitted on the corresponding training partition and then frozen.
Missing values are median-imputed before standardization.
The gate is shared across checkpoints, while \texttt{next\_checkpoint} and budget fraction identify the scheduled decision context.
At threshold $\tau$, the sequential policy is
\begin{equation}
\pi_\tau(s_{q,c}) =
\begin{cases}
\textsc{Continue}, & p_\theta(s_{q,c})\ge\tau,\\
\textsc{Stop}, & \text{otherwise}.
\end{cases}
\label{eq:policy}
\end{equation}
The policy never jumps directly to a predicted final budget.
This design is deliberate: if future reasoning is non-monotonic, the controller should re-observe newly generated evidence rather than commit to a one-shot budget prediction.

\subsection{Sequential Allocation and Calibration}

Inference begins at checkpoint 4.
For each still-active question, the system normalizes the currently available answers, computes \aec{} from the observed prefix, applies the frozen preprocessing statistics, and evaluates $p_\theta$.
If the probability is below $\tau$, the system returns the current Re$^2$ aggregate.
Otherwise it generates exactly the additional responses needed to reach the next scheduled checkpoint and repeats the procedure.
Checkpoint 128 is terminal regardless of gate output.
Therefore a trajectory visits checkpoints 4, 8, 16, 32, 64, and
128 in order, but it can never skip directly from checkpoint 4 to a
predicted terminal budget.
The online implementation batches active questions at each stage; stopped questions are removed from subsequent generation calls.

The gate score has no dataset-independent operating meaning, so deployment requires a separate calibration set.
Given candidate thresholds $\mathcal{T}$ and an allowed accuracy loss $\epsilon$, we select
\[
\begin{aligned}
\tau^\star &\in
\arg\min_{\tau\in\mathcal{T}}
\operatorname{Cost}_{\mathrm{cal}}(\pi_\tau),\\
\text{s.t.}\quad
\operatorname{Acc}_{\mathrm{cal}}(\pi_\tau)
&\geq
\operatorname{Acc}_{\mathrm{cal}}(\text{Fixed-128})-\epsilon.
\end{aligned}
\]
Ties are resolved before test access.
If the feasible set is empty, the safe action is to retain full compute rather than silently optimize a different utility.

\section{Experiments}

  We evaluate whether \method{} improves the accuracy--compute trade-off over
  fixed budgets and adaptive controls, and whether this advantage survives
  threshold selection without test access.
  Our evaluation combines complete offline frontiers, nested calibration, and
  incremental generation with a frozen threshold.

\subsection{Experimental Setup}

We control a frozen Qwen2.5-7B Re$^2$ reasoner \cite{wang2026re2}.
Responses are sampled at temperature 0.6 and top-$p=0.95$, with a maximum
generation length of 16,384 tokens.
For each problem, we allow at most 128 responses and expose cumulative prefixes
at checkpoints $\{4,8,16,32,64,128\}$.
We use one generation seed per protocol; exact seeds and prompts are reported
in the supplementary material.
At every checkpoint, answers are normalized and aggregated by the original
Re$^2$ evaluation rule.
Our primary benchmarks are GSM8K (1,319 test questions) \cite{cobbe2021gsm8k} and GPQA Diamond (198 questions) \cite{rein2024gpqa}.
\subsection{Metrics and Evaluation Protocols}

The primary evaluation is \emph{offline sequential replay}.
The complete response pool is generated before controller evaluation, but the replay policy reveals only the current checkpoint state when making each action.
Once the policy selects a stopping checkpoint, the original Re$^2$ evaluator
scores the corresponding raw response prefix using the same answer extraction,
redo filtering, and aggregation procedure as the full-compute system.
Policy actions depend only on the revealed prefix; canonical correctness remains
unavailable during replay.

We report Re$^2$ accuracy, mean responses per question, and response saving
relative to the 128-response budget.
For a policy stopping at $c_q$ responses on question $q$, response saving is
$100(1-\mathbb{E}_q[c_q]/128)$.
We report complete frontiers for $\tau\in\{0.1,\ldots,0.9\}$.
Mean responses measures sampling count rather than wall-clock latency or FLOPs;
completion-token savings are reported separately when available.

We additionally conduct frozen-threshold incremental-generation tests.
Calibration questions are generated to 128 responses, a threshold is selected using calibration data only, and that threshold is frozen before a disjoint test set is accessed.
On test questions, generation stops when the controller stops; a separately generated paired Fixed-128 pool supplies the reference.

\paragraph{Question-level splits.}
All checkpoints belonging to one question remain in the same fold.
In the ordinary five-fold analysis, feature preprocessing and optimization use the complementary questions, while held-out-fold loss determines early stopping before predictions on that fold are replayed.
This protocol is useful for descriptive frontiers but is not an untouched outer-test estimate.
Our nested analysis therefore divides each outer-training split again into model-training, inner-validation, and calibration partitions.
Inner validation selects training duration, calibration selects the operating threshold, and the frozen model--threshold pair is evaluated once on the outer fold.

\subsection{Baselines}

Because \method{} allocates cumulative response blocks, its controlled
comparisons use the same response pools, checkpoint schedule, and Re$^2$
aggregation rule.
We compare against the following controls.
\emph{Fixed budget} always stops at one of
$c\in\{4,8,16,32,64,128\}$.
\emph{Evidence heuristics} stop at the first checkpoint where majority ratio or
semantic consensus exceeds a threshold, or normalized entropy falls below one;
we display their complete post-hoc sweeps.
\emph{Matched random} shuffles \method{}'s selected checkpoints across
questions, exactly preserving its empirical budget distribution.
\emph{One-shot allocation} predicts a terminal checkpoint once from the
checkpoint-4 state and cannot revise that decision.
\emph{Oracle utility} selects checkpoints using future correctness and is a
non-deployable upper reference.
We additionally implement two direct response-sampling baselines on the same
fixed response pools.
\emph{Checkpoint-aligned Early-Stopping Self-Consistency (ESC)} applies the
published window-unanimity rule only at the checkpoints available to every
policy; its default window length is $w=5$.
\emph{Checkpoint-aligned Adaptive-Consistency (ASC)} applies the published
Beta-posterior stopping criterion at those checkpoints; its default confidence
parameter is $C=0.95$.
These aligned variants preserve the original decision statistics while sharing
\method{}'s action space and evaluator.
We do not label a response-only proxy as DSC because faithful DSC additionally
requires question-difficulty ranking and budget pre-allocation.

\subsection{Implementation Details}

The gate uses $\lambda=0.10$, five question-level folds, Adam optimization with learning rate $10^{-3}$, weight decay $10^{-4}$, up to 300 epochs, and patience 25.
Semantic features are derived from a frozen lightweight sentence encoder and
compressed to a 32-dimensional projection; the exact encoder and preprocessing
configuration are documented in the supplementary material for reproducibility.
The final controller uses 55 numerical inputs.
Only a predefined set of checkpoint-observable features is admitted to the
controller; outcome-like and future-derived fields are excluded.
The supplementary material gives the complete feature inventory and a
provenance audit of excluded columns.
Unless explicitly stated otherwise, heuristic sweeps and oracle comparisons are
diagnostic references rather than independently calibrated policies.

\subsection{Offline Accuracy--Compute Frontiers}

\begin{figure}[t]
\centering
\includegraphics[width=\columnwidth]{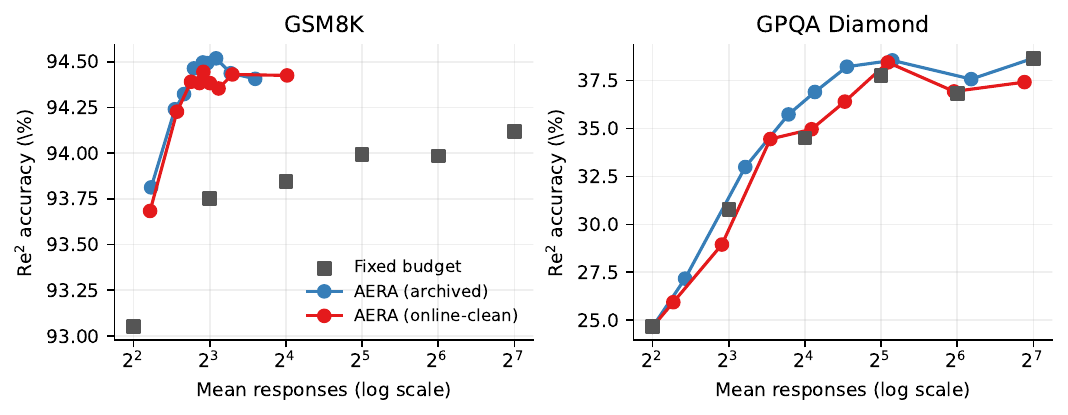}
\caption{Complete offline accuracy--response frontiers for fixed budgets,
checkpoint-aligned ESC and ASC, and \method{}.}
\label{fig:frontier}
\end{figure}

Figure~\ref{fig:frontier} is the primary summary of the complete offline
accuracy--response trade-off.
Table~\ref{tab:main} reports representative points from those measured
frontiers for readability.
These points are selected retrospectively to illustrate different trade-offs
and are descriptive rather than independently calibrated operating points.
Numerical sweeps appear in the supplementary material.

Checkpoint-aligned ESC is substantially more conservative: its standard
five-response unanimity window uses 20.30 responses on GSM8K and 115.80 on
GPQA, obtaining 94.31\% and 38.85\%, respectively.
Thus \method{} reaches a similar accuracy range with markedly fewer exposed
responses than this stability rule.
ASC is more competitive: at its published 0.95 confidence threshold it obtains
94.43\% at 9.40 responses on GSM8K and 38.99\% at 81.82 responses on GPQA.
Against nondominated ASC mixtures using no more computation than \method{} at
$\tau=0.3$, the paired differences are $-0.09$ points on GSM8K
(95\% CI $[-0.30,+0.09]$) and $-0.93$ on GPQA
($[-3.45,+1.46]$), establishing comparable rather than uniformly superior
aggregate accuracy.

\begin{table}[t]
\centering
\small
\begin{tabular}{llrr}
\toprule
Dataset & Method & Resp./q & Acc. (\%) \\
\midrule
GSM8K & Fixed-32 & 32.00 & 93.99\\
 & Fixed-128 & 128.00 & 94.12\\
 & ESC-aligned ($w=5$) & 20.30 & 94.31\\
 & ASC-aligned ($C=.95$) & 9.40 & 94.43\\
 & \method{} ($\tau=.3$) & 8.67 & 94.35\\
\midrule
GPQA & Fixed-32 & 32.00 & 37.75\\
 & Fixed-64 & 64.00 & 36.83\\
 & Fixed-128 & 128.00 & 38.65\\
 & ESC-aligned ($w=5$) & 115.80 & 38.85\\
 & ASC-aligned ($C=.95$) & 81.82 & 38.99\\
 & \method{} ($\tau=.3$) & 34.02 & 38.44\\
\bottomrule
\end{tabular}
\caption{Representative points on the descriptive, question-disjoint offline
accuracy--response frontiers. \method{} thresholds are selected
retrospectively from the complete sweep and are not confirmatory test settings.}
\label{tab:main}
\end{table}

\paragraph{Paired uncertainty.}
Question-level paired bootstrap with 20,000 resamples compares \method{}
decisions to Fixed-128.
At $\tau=0.3$, the accuracy difference is $+0.24$ points on GSM8K, with 95\% interval $[-0.03,+0.49]$, and $-0.21$ points on GPQA, with interval $[-3.34,+2.84]$.
Neither interval establishes an accuracy difference at the 5\% level.

\paragraph{Compute-matched routing.}
To isolate question-specific routing from the marginal budget distribution, we
randomly permute \method{}'s stopping checkpoints across questions within each
evaluation fold.
This preserves the complete checkpoint distribution and hence mean response
count, while removing dependence between a question and its assigned budget.
At $\tau=0.3$, \method{} exceeds this matched permutation by $1.18$ accuracy
points on GSM8K (94.35\% versus 93.17\% at 8.67 responses) and by $4.50$
points on GPQA (38.44\% versus 33.94\% at 34.02 responses); both one-sided
randomization tests give $p<10^{-4}$.
The gain therefore reflects question-specific routing rather than the stopping
distribution alone.

\paragraph{Nested calibrated replay.}
As a stricter complement to the descriptive frontiers, an outer question split is evaluated only after epoch and threshold selection on inner training, validation, and calibration partitions.
This protocol obtains 94.28\% accuracy at 7.88 responses on GSM8K, versus 94.12\% at Fixed-128 (paired difference $+0.17$ points, 95\% CI $[-0.17,+0.50]$).
The nested analysis therefore preserves the favorable GSM8K trade-off under
strict model and threshold selection; complete fold-level results appear in the
supplementary material.
Under the same two-point accuracy-constrained calibration rule, AERA and ASC
are indistinguishable on GSM8K (93.72\% versus 93.70\%, at 4.48 versus 4.22
responses).
The corresponding aggregate GPQA comparison favors ASC, indicating that the
learned gate is less stable under small-sample calibration shift; full
fold-level results are reported in the supplementary material.

\subsection{When Does Residual Gating Help?}

Aggregate comparisons can obscure the regime that motivates residual-utility
prediction.
We therefore conduct two hindsight diagnostics; neither policy observes the
future outcomes used in these analyses.
First, among questions that \method{} sends beyond checkpoint 4 at $\tau=0.3$,
42.4\% on GSM8K and 47.0\% on GPQA admit a higher-scoring later checkpoint,
compared with 0.9\% and 33.3\% among questions stopped at checkpoint 4.
The corresponding mean best-future gains are 14.14 versus 0.09 points on
GSM8K and 24.58 versus 1.78 points on GPQA.
Thus the continuation decisions are enriched for questions with residual
opportunity, although the grouping is retrospective.
Second, we group questions by their canonical correctness trajectory.
On the 16 GPQA questions containing a correct-to-incorrect collapse, the nested
out-of-fold 55-feature gate at $\tau=0.3$ obtains 69.24\% Re$^2$ accuracy
at 17.25 responses, whereas ASC at its default confidence obtains 36.98\% at
110.0 responses.
The paired difference is $+32.25$ points with a 95\% bootstrap interval
$[+15.89,+49.59]$.
On the 16 GSM8K collapse questions, the corresponding difference is $+2.91$
points with interval $[-0.53,+7.91]$, at 82.5 versus 92.5 responses.
These small subsets do not establish global superiority, but support the
narrower claim that prospective supervision is useful when apparent agreement
precedes degradation.

\subsection{Frozen-Threshold Incremental Generation}

For GSM8K, 50 calibration questions and 300 untouched test questions are fixed before generation.
The pre-specified rule chooses the least-token threshold whose calibration accuracy is within two percentage points of Fixed-128, with no fallback if the constraint is infeasible.
It selects $\tau=0.9$: calibration favors AERA by 0.65 points, while the untouched test yields 92.61\% versus 93.01\% for Fixed-128.
The paired difference is $-0.41$ points with 95\% bootstrap interval $[-1.79,+0.80]$.
AERA exposes 4.44 responses on average, saving 96.53\% of response count and 95.99\% of completion tokens (95\% CI $[94.96,96.61]$).
The calibration-to-test sign change illustrates why calibration performance must not be reported as final performance.
Against Fixed-4, AERA gains only $+0.13$ points (paired 95\% CI $[-0.21,+0.58]$) while using 0.44 additional responses on average.
It is likewise statistically unresolved against every fixed budget in Table~\ref{tab:online-gsm}.
Consequently, this online test validates the frozen calibration-to-generation protocol and shows that the selected policy is close to full-compute accuracy; it does \emph{not} by itself establish that adaptive routing outperforms the best fixed budget on this easy benchmark.

\begin{table}[t]
\centering
\small
\begin{tabular}{lrrr}
\toprule
Method & Resp./q & Acc. (\%) & Saving (\%)\\
\midrule
Fixed-4   & 4.00   & 92.47 & 96.88\\
Fixed-8   & 8.00   & 93.07 & 93.75\\
Fixed-16  & 16.00  & 92.59 & 87.50\\
Fixed-32  & 32.00  & 92.52 & 75.00\\
Fixed-64  & 64.00  & 92.87 & 50.00\\
Fixed-128 & 128.00 & 93.01 & 0.00\\
\method{} & 4.44 & 92.61 & 96.53\\
\bottomrule
\end{tabular}
\caption{Frozen-threshold online GSM8K test on 300 untouched questions. The threshold is selected on a separate 50-question calibration set. AERA's separately measured completion-token saving is 95.99\%.}
\label{tab:online-gsm}
\end{table}

\paragraph{Question-level behavior.}
The online policy stops at four responses on 283 of 300 questions.
Its score is higher than Fixed-128 on 58 questions, lower on 15, and identical
on 227; the small aggregate loss is concentrated in a few routed cases rather
than explained by a broad decline.

\subsection{Ablation Study}

The one-shot predictor collapses toward checkpoint 4 under severe target
imbalance and is substantially less accurate, supporting sequential
re-observation.
The hindsight oracle shows remaining headroom, especially on GPQA.
Feature-group diagnostics, semantic-feature ablations, computation-cost
sensitivity, and the complete leakage audit are reported in the supplementary
material.

\section{Limitations and Broader Impact}
\label{sec:limitations}

Most results are obtained by replaying one frozen reasoner's response pools.
The online studies use one generation seed and research code, and therefore do
not establish latency, energy, throughput, or production‑overhead savings.
Offline response count is a sampling proxy; online completion‑token counts omit
prompt processing, semantic encoding, and controller overhead.

The controllers are trained within each benchmark.
Nested folds provide the cleanest within‑dataset estimate, while source‑to‑target
transfer is strongly asymmetric and does not support a universal gate.
GPQA and the AMC/AIME/MMLU‑Pro studies are also small, making modest
accuracy differences uncertain.
Representative ordinary‑replay thresholds are post hoc; only nested and online
results use held‑out calibration.
A direct ASC comparison is competitive on GSM8K and stronger in the aggregate
GPQA accuracy‑constrained evaluation, so our evidence supports a
collapse‑specific advantage rather than uniform dominance over posterior
stopping rules.
A separate GPQA online stress test did not satisfy its calibration constraint;
we therefore restrict the strict online‑efficiency claim to GSM8K and report the
full GPQA protocol in the supplementary material.
Nevertheless, GPQA illustrates why knowledge‑intensive scientific reasoning,
where candidate answers may be revised as evidence accumulates, is a demanding
setting for prospective rather than confidence‑only stopping criteria.

The unsupervised semantic projection is benchmark dependent.
Future work should evaluate multiple reasoners and seeds, learn conservative
calibration under distribution shift \cite{li2025unified}, and measure end‑to‑end serving cost.
Premature stopping can disproportionately harm difficult or shifted queries, so
high‑stakes systems should retain minimum budgets and support escalation.

\section{Conclusion}

Adaptive reasoning should predict whether future computation is useful,
rather than equating confidence with correctness.
\method{} implements this principle by learning recoverability from observable
response‑prefix evidence and re‑evaluating decisions after each allocated
response block.
Trajectory analysis, nested replay, and frozen online generation show this
separation delivers near‑full‑compute GSM8K accuracy with far lower
generation cost.
Compute‑matched randomization further shows gains stem from
question‑specific allocation, not marginal stopping distribution.
These findings make residual‑utility prediction a complement to
confidence‑based stopping when checkpoint correctness evolves non‑monotonically.
Together, results establish a clear decision principle and
task‑dependent controller, outlining steps toward broader adoption.

\bibliography{references}

\end{document}